\documentclass[10pt, a4paper]{article}

\usepackage{amsmath}
\usepackage{booktabs}
\usepackage{graphicx}  
\usepackage{listings}

\usepackage{tcolorbox}
\usepackage{array}
\usepackage[final]{lrec2026} 

\title{Identifying Imaging Follow-Up in Radiology Reports: \\A Comparative Analysis of Traditional ML and LLM Approaches}

\name{ \parbox{\textwidth}{\fontsize{12}{0}\selectfont\centering
Namu Park\textsuperscript{1*}\footnotemark[1],
Giridhar Kaushik Ramachandran\textsuperscript{2*}\footnotemark[2],
Kevin Lybarger\textsuperscript{2}\footnotemark[2] \\
Fei Xia\textsuperscript{3}\footnotemark[3],
Özlem Uzuner\textsuperscript{2}\footnotemark[2],
Meliha Yetisgen\textsuperscript{1}\footnotemark[1],
Martin Gunn\textsuperscript{4}\footnotemark[4]
}}

\address{
$^{1}$Department of Biomedical Informatics and Medical Education, School of Medicine, \\University of Washington, Seattle, WA, USA\\
$^{2}$Department of Information Sciences and Technology, George Mason University, Fairfax, VA, USA\\
$^{3}$Department of Linguistics, University of Washington, Seattle, WA, USA\\
$^{4}$Department of Radiology, School of Medicine, University of Washington, Seattle, WA, USA\\[4pt]
*The first two authors contributed equally to this work
}

\abstract{
Large language models (LLMs) have shown considerable promise in clinical natural language processing, yet few domain-specific datasets exist to rigorously evaluate their performance on radiology tasks. In this work, we introduce an annotated corpus of 6,393 radiology reports from 586 patients, each labeled for follow-up imaging status, to support the development and benchmarking of follow-up adherence detection systems. Using this corpus, we systematically compared traditional machine-learning classifiers—logistic regression (LR), support vector machines (SVM), Longformer, and a fully fine-tuned Llama3-8B-Instruct—with recent generative LLMs. To evaluate generative LLMs, we tested GPT-4o and the open-source GPT-OSS-20B under two configurations: a baseline (Base) and a task-optimized (Advanced) setting that focused inputs on metadata, recommendation sentences, and their surrounding context. A refined prompt for GPT-OSS-20B further improved reasoning accuracy. Performance was assessed using precision, recall, and F1 scores with 95\% confidence intervals estimated via non-parametric bootstrapping. Inter-annotator agreement was high (F1 = 0.846). GPT-4o (Advanced) achieved the best performance (F1 = 0.832), followed closely by GPT-OSS-20B (Advanced; F1 = 0.828). LR and SVM also performed strongly (F1 = 0.776 and 0.775), underscoring that while LLMs approach human-level agreement through prompt optimization, interpretable and resource-efficient models remain valuable baselines.
\\ \newline  }

\begin{document}
\pagenumbering{gobble}
\maketitleabstract

\section{Introduction}
Follow-up recommendations are frequently included in radiology reports, varying depending on the modality and body region, with an average of 19.9\% of reports containing such recommendations \cite{lau2020extraction}. These often, but not always, specify an imaging modality, reason, and timeframe. The referrer is typically responsible for ensuring patient communication and follow-up \cite{larson_actionable_2014}, but radiology departments usually do not track recommendations and may be unaware if follow-ups are ordered or completed. Accordingly, follow-up adherence is suboptimal, with over a third of recommendations not being followed up as reported in a recent publication \cite{mabotuwana2019automated}. Although undoubtedly a sizeable proportion of follow-up imaging is not ordered because it is felt unnecessary by the patient or clinician, one study found that more than 35\% of recommendations were not followed up simply due to the referring clinician not acknowledging them \cite{callen_failure_2012}. Reasons related to failure to follow-up, especially on incidental findings, include referring clinician missing the recommendations or losing track of them while addressing a more acute illness, loss of information during handover between care teams, the recommendation not being communicated to the patient, and the patient failing to schedule or show up for the follow-up appointment\cite{kulon_lost_2016}. 

Missing recommended follow-ups can have serious clinical consequences, particularly when incidental findings represent early manifestations of malignancy or other progressive disease. Beyond individual patient harm, inadequate follow-up adherence also contributes to inefficiencies in healthcare delivery and increased system costs \cite{kadom2022novel}. Accordingly, improving the identification and tracking of follow-up recommendations is essential for ensuring timely care and reducing preventable morbidity and mortality. However, progress in this area has been limited in part by the lack of publicly available, well-annotated corpora tailored to this clinical problem.

To address this gap, we developed a corpus of 6,393 radiology reports from 586 patients, each manually annotated for the follow-up imaging status. Using this resource, we systematically evaluated multiple machine learning and large language model (LLM) approaches for the task of follow-up identification in free-text radiology reports—a domain characterized by high linguistic variability, implicit clinical reasoning, and long-context dependencies. The evaluated models included logistic regression (LR), support vector machines (SVM)\cite{hearst1998support}, Longformer \cite{beltagy2020longformer}, and a fully fine-tuned Llama3-8B-Instruct \cite{llama3}, alongside generative models GPT-4o\cite{hurst2024gpt} and GPT-OSS-20B. By introducing this annotated corpus and benchmarking across both traditional and generative paradigms, this study provides a valuable resource and empirical foundation for advancing real-world clinical NLP applications in radiology.

\section{Related Work}

Given the critical importance of identifying follow-up recommendations to prevent potentially fatal outcomes, numerous NLP-based methods have been developed for radiology reports. For instance, Yetisgen et al. introduced a supervised text classification approach to detect recommendation sentences, leveraging features beyond simple unigram tokens \cite{yetisgen2011automatic}. Conditional Random Fields (CRF) \cite{lafferty2001conditional} have also been utilized to capture temporal information within follow-up recommendations \cite{xu2012named}. 

Other studies have compared traditional machine learning models, such as Support Vector Machines (SVMs) \cite{hearst1998support} and Random Forests (RFs) \cite{breiman2001random}, with deep learning approaches like Recurrent Neural Networks (RNNs) \cite{schuster1997bidirectional} for identifying follow-up recommendations \cite{carrodeguas2019use}. Lau et al. created a corpus of radiology reports annotated with entities related to follow-up recommendations, trained a neural model on this dataset, and applied it at scale to assess adherence rates \cite{lau2020extraction}. 

In addition, other research has focused on building tools that evaluate follow-up compliance by extracting essential elements such as the recommended time frame and imaging modality from radiology reports \cite{mabotuwana_improving_2018}. Dalal et al. tackled the challenge of tracking follow-up adherence with an Extremely Randomized Trees model \cite{geurts2006extremely} that integrates various clinical features, including recommendation attributes, metadata, and text similarity \cite{dalal_determining_2020}. This model achieved an F1 score of 0.807, closely approaching the inter-annotator agreement of 0.853 F1, demonstrating its effectiveness in tracking follow-up adherence and its potential utility in real-world clinical settings. However, the application of generative LLMs to this specific task is still understudied. Therefore, using our annotated corpus as a benchmark resource, we aim to evaluate both traditional machine learning models and generative LLMs for the task of follow-up identification, providing insights into model performance, generalizability, and practical suitability for clinical deployment.

\section{Methods}
This retrospective study was approved by the Institutional Review Board (IRB) at the authors’ institution and satisfies the waiver of patient informed consent. All radiology reports were de-identified prior to annotation. Python code used for the experiments is available on our GitHub repository: \url{https://github.com/XXX}.

\subsection{Task Description}
\begin{figure}[ht]
\centerline{\includegraphics[width=\columnwidth]{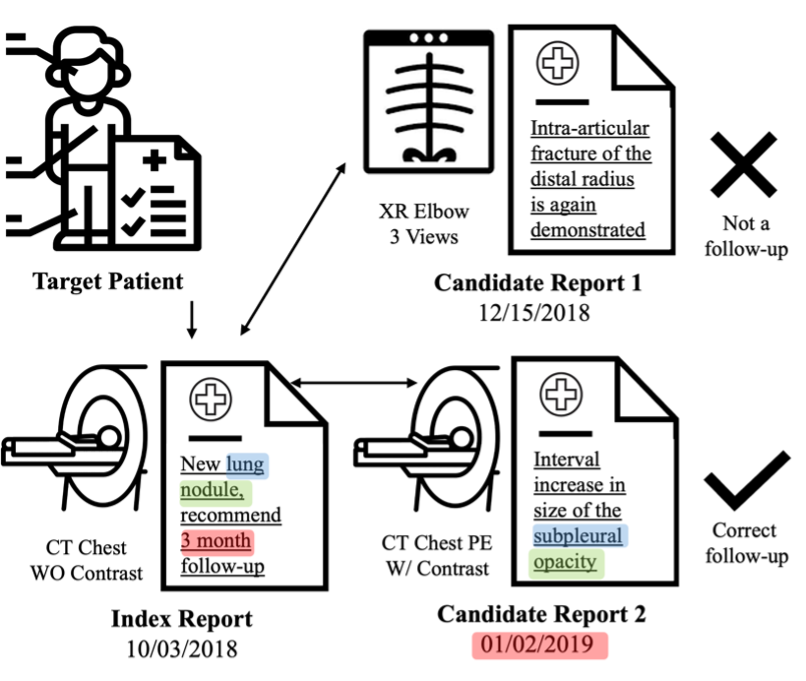}}
\caption{Follow-up identification task}
\label{aba:fig1}
\end{figure}

To track completion of follow-up imaging tests, we define two types of radiology reports. Reports that contained a finding for which an imaging follow-up recommendation was present were termed “index reports.” Follow-up recommendations are often associated with potential malignancy and typically (though not invariably) specify the recommended follow-up timeframe, imaging modality, and anatomy or pathology to be re-imaged (e.g., “recommend follow-up with chest CT in 6 months to assess stability of the lung nodule”). For a given index report, all subsequent imaging reports for the same patient in the radiology information system (RIS) were considered—referred to hereafter as “candidate reports.” Thus, each patient is represented by a single index report and its associated candidate reports. Figure~\ref{aba:fig1} provides a visual overview of the follow-up identification task.

\subsection{Dataset}
\subsubsection{Sampling Process}
We used a clinical database containing 7 million radiology reports from 959{,}382 patients, covering the years 2007–2020. All reports were de-identified using a neural de-identifier \cite{leeDobbins2021}. During this period, each patient underwent an average of 7.45 radiology examinations, with 29.9\% having only a single report.

\begin{figure*}[htb!]
  \begin{center}
    \includegraphics[width=6.0in]{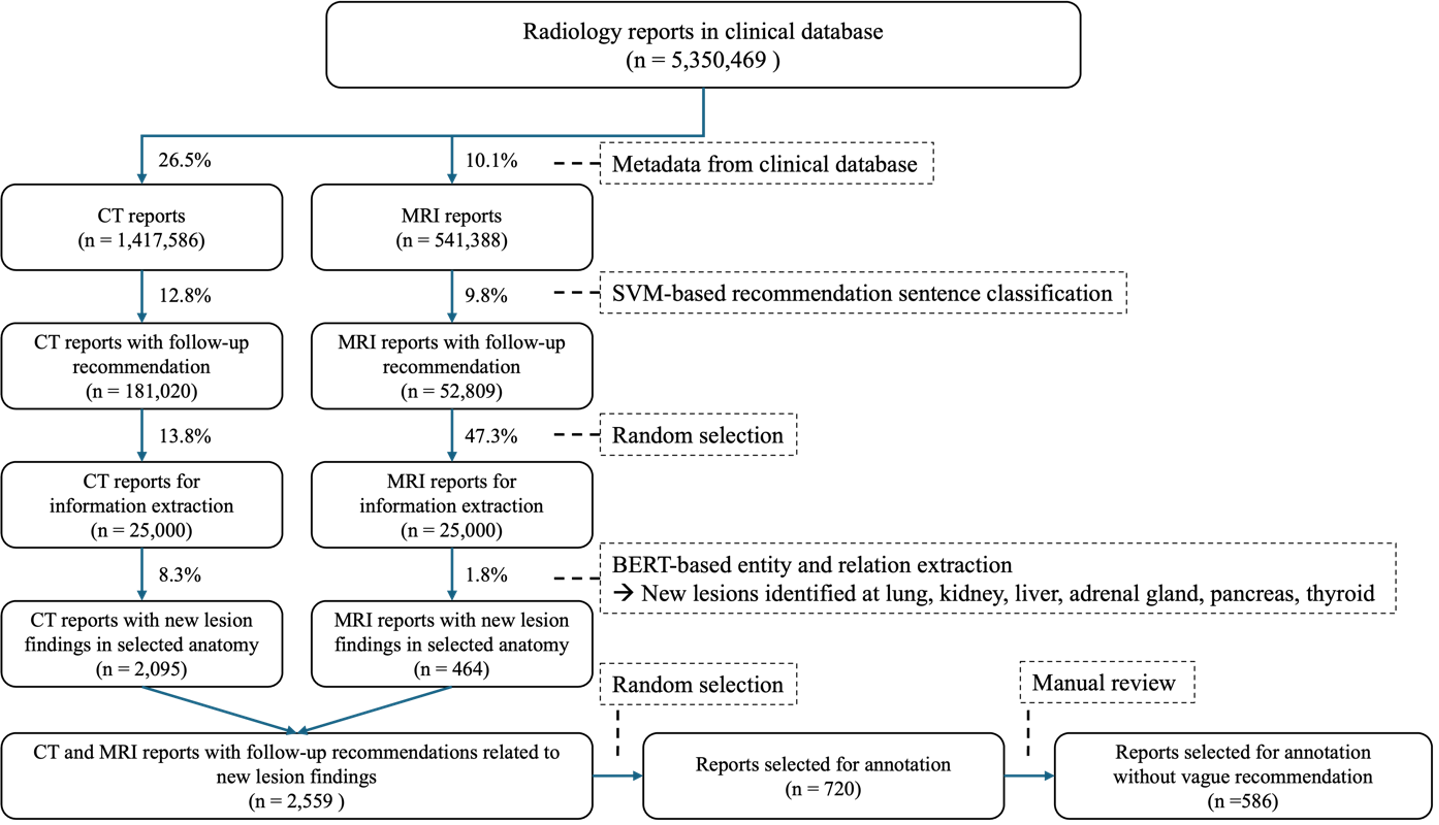}
  \end{center}
  \caption{Sampling process for index reports}
  \label{aba:fig2}
\end{figure*}

We focused on selecting the most relevant index reports for our task using the approaches described in Figure~\ref{aba:fig2}. We limited index report modality to CT and MRI, as these frequently include follow-up recommendations.

To ensure rapid and accurate sampling, we combined automated and manual methods. First, to increase the prevalence of recommendation sentences among candidate index reports, we developed a lightweight SVM-based recommendation sentence classifier using an existing annotated corpus \cite{lau2022event}. Trained on binary sentence labels (\textit{has recommendation} vs.\ \textit{no recommendation}), the classifier evaluated each sentence and marked it as a recommendation when appropriate. The sentence-level performance was 0.87 F1, and it identified 181{,}020 CT and 52{,}809 MRI reports with at least one recommendation sentence—12.8\% and 9.8\% of notes per modality, respectively.

From reports with at least one follow-up recommendation, we next identified those including a mass lesion. After randomly selecting 25{,}000 reports per modality, we applied a BERT-based entity and relation extractor from prior work \cite{park2024novel} to extract detailed clinical findings (mass lesions, problems, attributes). We targeted cases with newly discovered lesion findings in six anatomies—lung, kidney, liver, adrenal gland, pancreas, and thyroid—yielding 2{,}095 CT reports and 464 MRI reports, each with at least one new lesion in a target anatomy and at least one recommendation sentence.

From this refined pool, 720 index reports were randomly selected. Two medical student annotators then excluded potentially incorrect samples from automated filtering, removing reports with excessively vague recommendations (e.g., “\textit{attention on follow-up is recommended},” “\textit{clinical correlation is recommended}”) and those recommending or leading to same-day characterization of acute findings. The combined filtering resulted in 586 index reports, with 134 inappropriate samples removed. For each of the 586 index reports, corresponding candidate reports were sampled without restriction on imaging modality, resulting in 5{,}807 candidate reports (mean 9.9 per index; range [1, 168]).

\subsubsection{Data Annotation}
All reports were annotated by two medical students, who identified the earliest qualifying follow-up report among the candidate reports, if it existed. A candidate report was considered a follow-up if it addressed the same anatomical region as the lesion in the index report and explicitly referenced or negated the prior finding (e.g., \textit{“multiple pulmonary nodules unchanged from the previous examination,” “redemonstration of hypodense mass measuring 2.6 $\times$ 2.4 cm, previously 3.2 $\times$ 3.0 cm”}). In multi-lesion cases, identifying the most suitable follow-up exam was challenging; such complex cases were flagged and reviewed with a board-certified radiologist. The annotation process comprised 16 rounds. In the first 11 rounds, samples were doubly annotated, disagreements were resolved in weekly meetings, and any controversial cases were adjudicated with a board-certified radiologist. After 11 rounds, inter-annotator agreement was 0.846 F1. The subsequent 5 rounds were single annotated to increase volume, resulting in 347 single-annotated report pairs and 239 double-annotated report pairs.

Among the 586 index–candidate pairs, a qualifying follow-up report was identified in 417 pairs (71.3\%), while 169 index reports (28.7\%) had no follow-up identified. For 54\% of index reports, the first or second candidate chronologically was labeled as the follow-up. While index reports included only CT and MRI, candidate reports spanned modalities: computed/digital radiography (n=2{,}139, 36.8\%), CT (n=2{,}003, 34.5\%), MRI (n=520, 8.9\%), ultrasound (n=516, 8.8\%), nuclear medicine (n=171, 2.9\%), angiography (n=127, 2.2\%), PET-CT (n=99, 1.7\%), mammography (n=65, 1.1\%), etc. Index reports averaged 437.5 tokens (30.4 sentences) versus 252 tokens (17.3 sentences) for candidate reports. We used the BERT tokenizer \cite{devlin2019bert}, which converts text into subword tokens.

\subsection{Models}
Prior work \cite{mabotuwana_improving_2018} in follow-up imaging identification applied discrete machine learning models combining text and engineered linguistic features to predict the likelihood of a correct candidate report. We framed follow-up identification as a binary classification task by constructing index–candidate pairs and labeling each pair positive if the candidate report was the correct follow-up for the index report.

\subsubsection{Feature-based Traditional Models}
\begin{figure}[!ht]
\begin{center}
\includegraphics[width=\columnwidth]{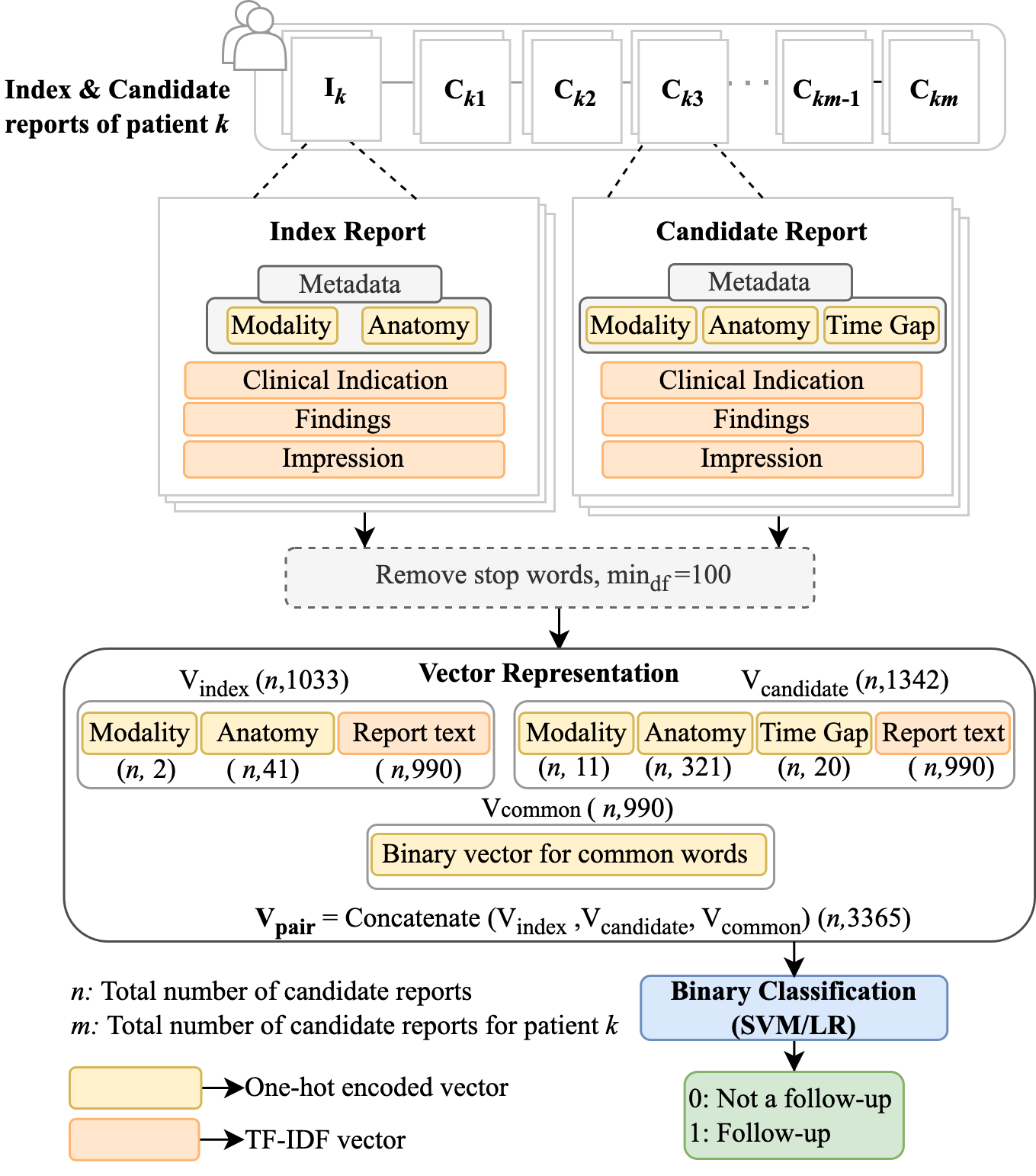}
\caption{Follow-up identification using feature-based traditional models -- SVM and LR}
\label{aba:fig3}
\end{center}
\end{figure}

We evaluated Logistic Regression (LR) and Support Vector Machines (SVM) in a supervised learning setting, with inputs consisting of index and candidate report text plus metadata (imaging modality and report time gap in days). Separate vector representations for index and candidate reports were created using TF-IDF \cite{sparck1972statistical}. Metadata were encoded using binary indicator vectors. A binary vector representing words shared by index and candidate reports was concatenated to form the final representation vector. The SVM used a sigmoid kernel and L2 regularization; both SVM and LR used class-balanced loss functions. Figure~\ref{aba:fig3} illustrates the SVM and LR pipelines. Individual vectors with separate encodings for metadata and text represented each of the index (\(V_{\text{index}}\)) and candidate (\(V_{\text{candidate}}\)) reports; the final pair vector (\(V_{\text{pair}}\)) concatenated index, candidate, and shared-word vectors.

\subsubsection{Supervised Learning using Transformer-based Models}
\begin{figure}[!ht]
\begin{center}
\includegraphics[width=\columnwidth]{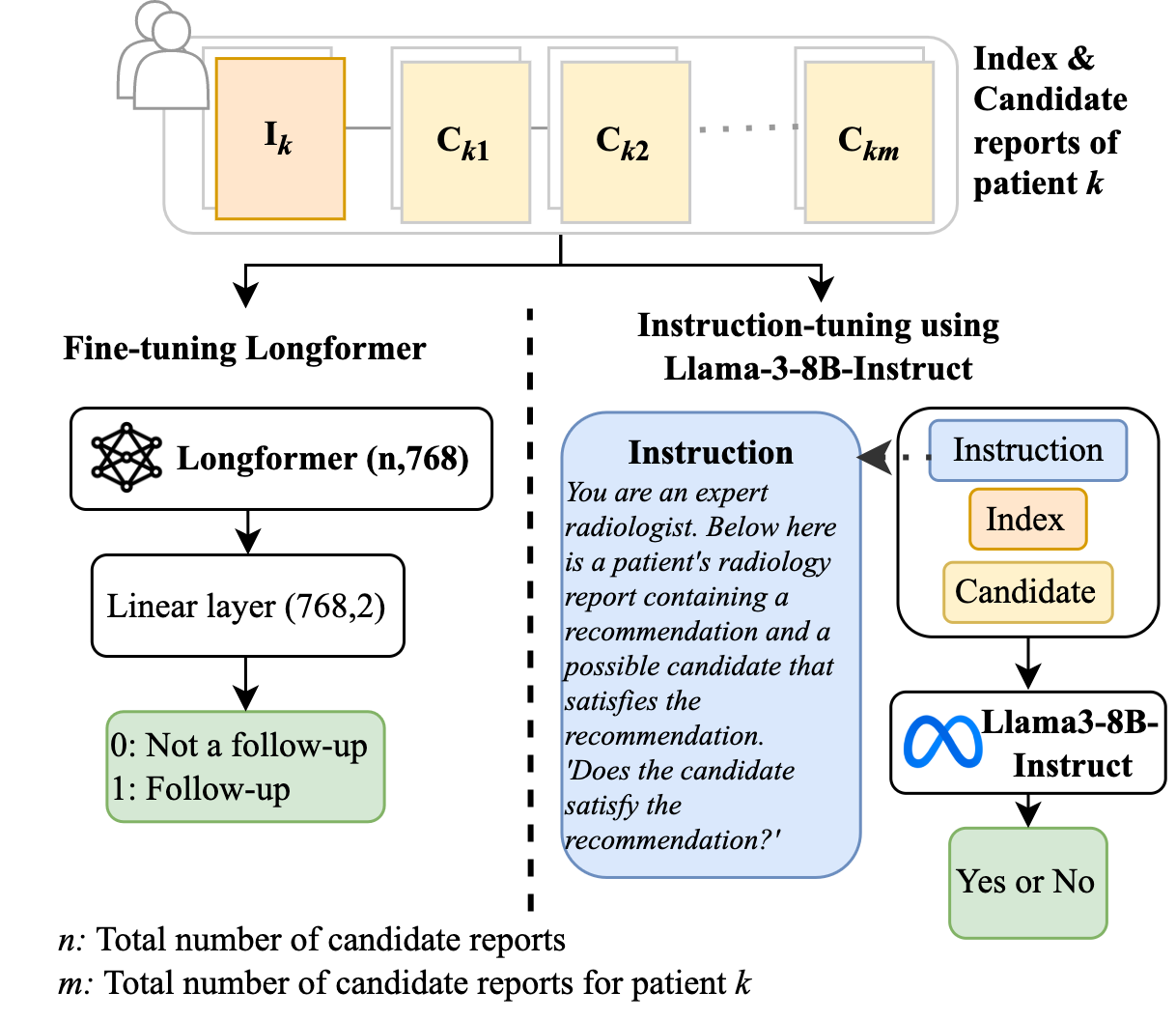}
\caption{Follow-up identification using Transformer-based models -- Longformer and Llama3-8B-Instruct}
\label{aba:fig4}
\end{center}
\end{figure}

We also investigated supervised learning with Transformer-based models capable of extended inputs, including Longformer \cite{beltagy2020longformer}, BioClinicalModernBERT \cite{sounack2025bioclinical}, Llama3-8B-Instruct \cite{llama3}, and Llama3.1-8B-Instruct \cite{grattafiori2024llama}, with model selection guided by compute constraints. We report results for Longformer and Llama3-8B-Instruct, which demonstrated the highest performance. Each input concatenated index and candidate report text separated by special tokens. Longformer used a linear classification layer. Both Longformer and Llama3-8B-Instruct were trained on the complete index and candidate text plus metadata. For Llama3-8B-Instruct, we tested several prompt designs; the best included a brief instructional prefix preceding the report pair (Figure~\ref{aba:fig4}). The model was then instruction-tuned via full supervised fine-tuning (SFT) using this engineered prompt. Additional implementation details are available in our GitHub repository.

\begin{figure*}[htb!]
  \begin{center}
    \includegraphics[width=6.0in]{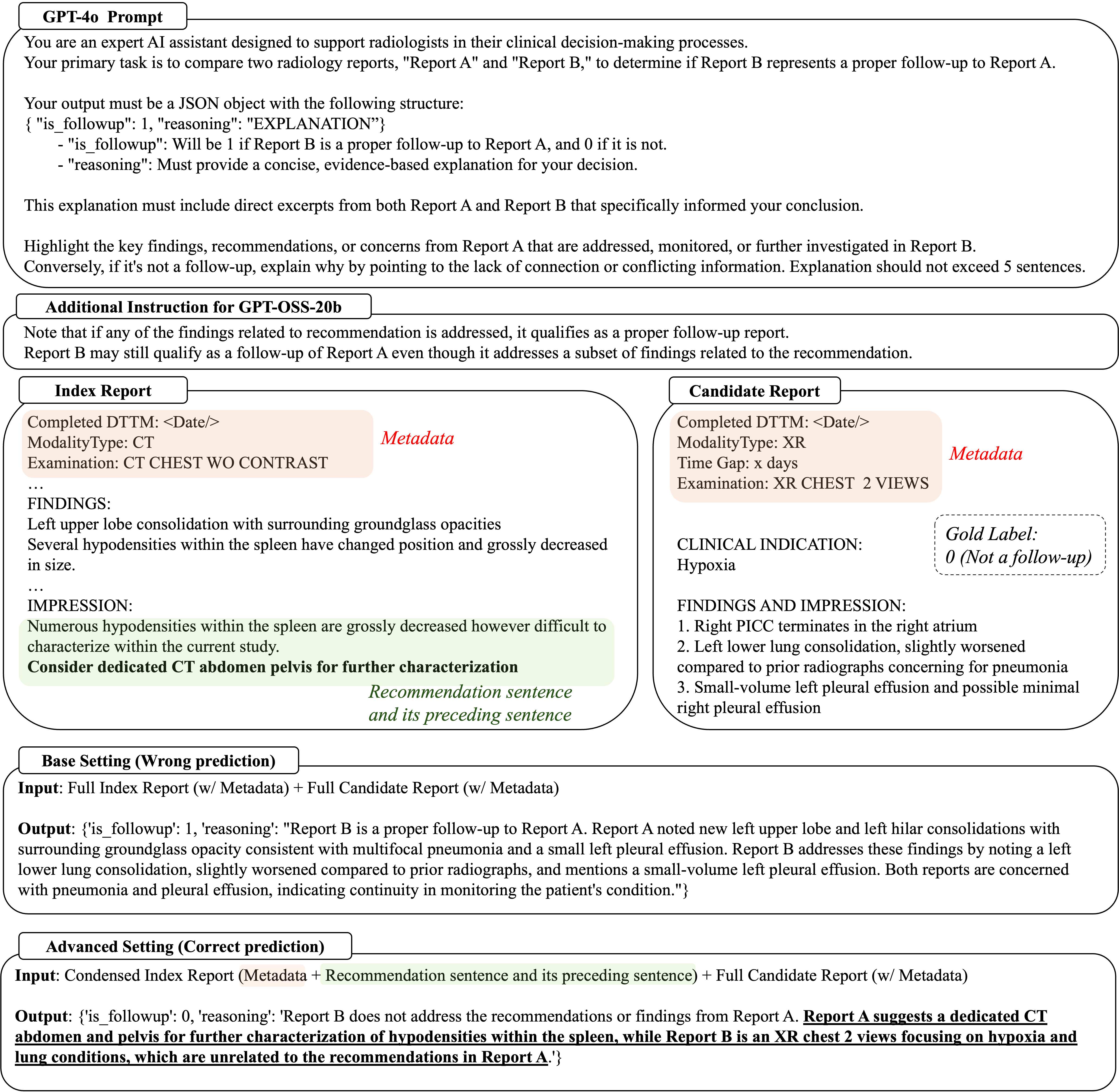}
  \end{center}
  \caption{Follow-up identification using generative LLMs. GPT-OSS-20B prompt is created by adding 2-sentence instructions on top of GPT-4o prompt. Outputs for base and advanced settings are actual predictions from GPT-4o.}
  \label{aba:fig5}
\end{figure*}

\subsection{Generative Large Language Model}
To evaluate generative LLMs for identifying follow-up imaging studies, we assessed GPT-4o \cite{hurst2024gpt} (version 2024-05-13) and GPT-OSS-20B \cite{agarwal2025gpt} in our HIPAA-compliant environment using two strategies: a baseline setting and an advanced, task-optimized setting (Figure~\ref{aba:fig5}). GPT-4o provides strong instruction-following and clinical reasoning, and GPT-OSS-20B is a recent open-source model suitable for secure institutional deployment. Larger variants such as GPT-OSS-120B were not used, as they offered only marginal gains in our task at substantially higher computational cost.

In the baseline setting, the model received the full index and candidate reports including metadata, accompanied by minimal task-specific instruction, and was asked to determine whether the candidate report represented an appropriate follow-up. In contrast, the advanced setting restricted input to metadata and the recommendation sentence—identified using the SVM-based sentence classifier in Section~3.2.1—along with its immediate preceding sentence (green box in Figure~\ref{aba:fig5}), emphasizing clinically relevant content for follow-up determination.

To optimize prompt engineering, we randomly selected 60 index reports and their corresponding candidate reports for iterative refinement during development. All prompt optimization was conducted using GPT-4o, and the final prompt that achieved the best development performance was adopted as the standard prompt for both the baseline and advanced settings. This optimized prompt was then directly applied to GPT-OSS-20B to assess cross-model transferability. The 60 development samples used for prompt optimization were excluded from final evaluation, which followed the performance criteria described in the next section.

Following initial error inspection on the development set, we observed that GPT-OSS-20B tended to judge candidate reports as incorrect when not all findings in the recommendation were re-addressed. To account for this behavior, we introduced an additional instruction on top of the GPT-4o prompt clarifying that a follow-up remains valid even if only a subset of the recommended findings is addressed (Figure~\ref{aba:fig5}).

\subsection{Evaluation}

\begin{figure*}[htb!]
  \begin{center}
    \includegraphics[width=6.0in]{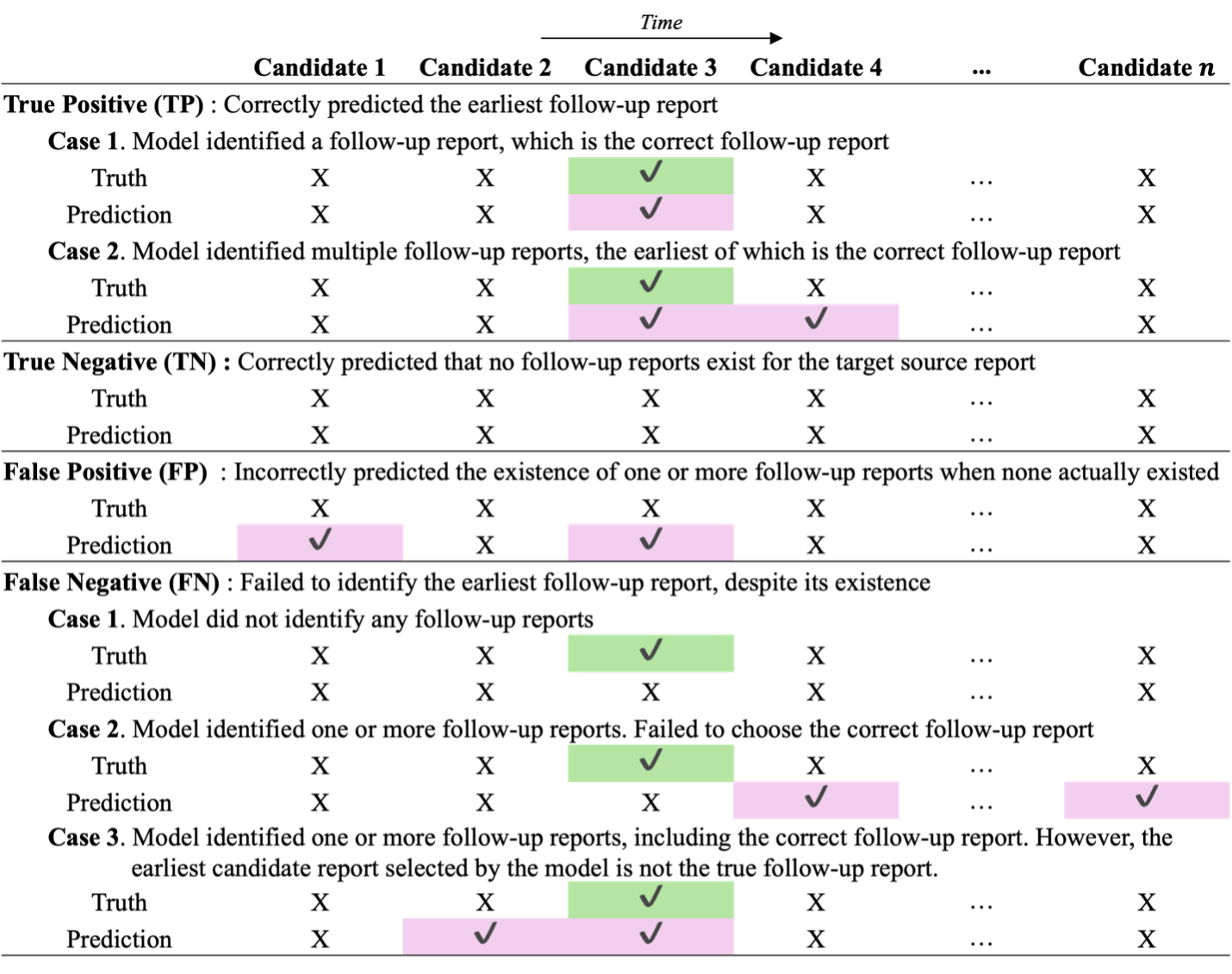}
  \end{center}
  \caption{Evaluation method for follow-up identification task. The check mark indicates that the target candidate report has been identified as a follow-up report either by the model (Prediction) or by the annotators (Truth). Boxes in purple show the prediction by the model and boxes in green are the true labels}
  \label{aba:fig6}
\end{figure*}

We evaluated performance at the index report-level chronologically (Figure \ref{aba:fig6}) using the following categories: 1) True Positive (TP) - correctly predicted the follow-up candidate report, if it existed. If multiple positive predictions existed, only the first one was considered for comparison because this is the clinically most important examination to ensure that follow-up did occur. 2) False Positive (FP) - incorrectly identified a follow-up when it was not a follow-up; 3) False Negative (FN) - did not predict the correct follow-up which existed; and 4) True Negative (TN) - correctly predicted the absence of a follow-up when there was no matching follow-up examination. 

The evaluation set consisted of 526 index reports and their corresponding candidate reports, excluding 60 index reports and their associated candidate reports that were used for prompt tuning of GPT-4o. We evaluated our models in two settings: (1) five-fold cross-validation (CV) for supervised learning models and (2) in-context learning for generative LLMs. For CV, we set train-validation-test splits and tune the hyperparameters at every fold. We reported precision, sensitivity, F1, and specificity for all our approaches, along with their 95\% Confidence Intervals (CI). We utilized non-parametric bootstrap test to compare the models’ F1 scores with a significance threshold of 0.05.

\section{Results}

\begin{table*}[htb!]
\small
\centering
\caption{Model performance metrics with 95\% confidence intervals}
\renewcommand{\arraystretch}{0.9}
\begin{tabular}{lccc}
\toprule
\textbf{Model} & \textbf{Precision} & \textbf{Recall} & \textbf{F1} \\
\midrule
SVM & 0.769 (0.727, 0.812) & 0.785 (0.743, 0.827) & 0.777 (0.743, 0.809) \\
LR & 0.792 (0.750, 0.833) & 0.759 (0.715, 0.802) & 0.775 (0.740, 0.808) \\
Longformer & 0.741 (0.694, 0.788) & 0.639 (0.590, 0.687) & 0.686 (0.646, 0.724) \\
Llama3-8B-Instruct & \textbf{0.907 (0.872, 0.943)} & 0.623 (0.573, 0.670) & 0.738 (0.698, 0.775) \\
GPT-4o (Base) & 0.740 (0.698, 0.783) & 0.807 (0.767, 0.846) & 0.772 (0.738, 0.803) \\
GPT-4o (Advanced) & 0.841 (0.803, 0.878) & 0.825 (0.786, 0.863) & \textbf{0.832 (0.802, 0.860)} \\
GPT-OSS-20B (Base) & 0.734 (0.690,0.777) & 0.775 (0.733,0.818) & 0.753 (0.720,0.787) \\
GPT-OSS-20B (Advanced) & 0.824 (0.786,0.862) & \textbf{0.833 (0.795,0.870)} & 0.828 (0.799,0.856) \\
\bottomrule
\end{tabular}%
\label{aba:tbl1}
\end{table*}

Table~\ref{aba:tbl1} summarizes the aggregated performance of all evaluated models. Among them, GPT-4o (Advanced) achieved the highest overall performance, with an F1 score of 0.832 (95\% CI: 0.802–0.860)—closely approximating the inter-annotator agreement score of 0.846. The next best performer was GPT-OSS-20B (Advanced), which achieved an F1 of 0.828 (0.799–0.856), demonstrating nearly equivalent performance to GPT-4o despite being a smaller, fully open-source model. Both advanced configurations markedly outperformed their respective base counterparts, underscoring the benefit of the optimized input design that emphasized follow-up recommendation cues within the radiology reports. Specifically, GPT-4o (Advanced) achieved a precision gain of $\Delta$+0.101, a recall increase of $\Delta$+0.018, and a corresponding F1 improvement of $\Delta$+0.060 relative to its base setting. Similarly, GPT-OSS-20B (Advanced) improved precision by $\Delta$+0.090, recall by $\Delta$+0.058, and F1 by $\Delta$+0.075 compared to its base configuration. Statistical significance testing (Table~\ref{aba:tbl2}) confirmed that these gains were significant (p~\textless~0.05) for each model relative to their baselines. However, no statistically significant difference was observed between GPT-4o (Advanced) and GPT-OSS-20B (Advanced), suggesting that GPT-OSS-20B can serve as a viable and complementary alternative to GPT-4o—particularly in resource-constrained or privacy-sensitive environments where closed-source APIs are less feasible.

The fully fine-tuned Llama3-8B-Instruct model achieved the highest precision among all systems (0.907, 95\% CI: 0.872–0.943), yet its recall remained substantially lower (0.623, 0.573–0.670), resulting in a reduced F1 score compared to most other approaches. This imbalance may reflect overfitting during fine-tuning, where the model captured training-specific patterns at the expense of generalization to unseen cases. Alternatively, the limited recall could stem from the pronounced class imbalance in the dataset—only 417 candidate reports (7.2\% of total pairs) represented true follow-ups. Such skewed distributions can bias models toward negative predictions, diminishing sensitivity to the minority class. Future work could explore class-balanced sampling or focal loss optimization to mitigate this effect and improve model robustness on rare follow-up cases.

\begin{table*}[]
\centering
\caption{Pairwise statistical significance comparisons for our models using the non-parametric bootstrap test (n=253 patients drawn with replacement, 10,000 repetitions). P-values are indicated in parentheses.}
\label{aba:tbl2}
\resizebox{\textwidth}{!}{%
\begin{tabular}{c|cccccccc}
\hline
\multicolumn{1}{l|}{} &
  \begin{tabular}[c]{@{}c@{}}GPT-4o\\ (Advanced)\end{tabular} &
  \begin{tabular}[c]{@{}c@{}}GPT-OSS-20B\\ (Advanced)\end{tabular} &
  \begin{tabular}[c]{@{}c@{}}GPT-4o\\ (Base)\end{tabular} &
  \begin{tabular}[c]{@{}c@{}}GPT-OSS-20B\\ (Base)\end{tabular} &
  SVM &
  LR &
  Llama3-8B &
  Longformer \\ \hline
\begin{tabular}[c]{@{}c@{}}F1\\ (95\% CI)\end{tabular} &
  \begin{tabular}[c]{@{}c@{}}0.832\\ (0.802, 0.860)\end{tabular} &
  \begin{tabular}[c]{@{}c@{}}0.828\\ (0.799, 0.856)\end{tabular} &
  \begin{tabular}[c]{@{}c@{}}0.772\\ (0.738, 0.803)\end{tabular} &
  \begin{tabular}[c]{@{}c@{}}0.753\\ (0.720, 0.787)\end{tabular} &
  \begin{tabular}[c]{@{}c@{}}0.777\\ (0.743, 0.809)\end{tabular} &
  \begin{tabular}[c]{@{}c@{}}0.775\\ (0.740, 0.808)\end{tabular} &
  \begin{tabular}[c]{@{}c@{}}0.738\\ (0.698, 0.775)\end{tabular} &
  \begin{tabular}[c]{@{}c@{}}0.686\\ (0.646, 0.724)\end{tabular} \\ \hline
\begin{tabular}[c]{@{}c@{}}GPT-4o\\ (Advanced)\end{tabular} &
  N/A &
  \begin{tabular}[c]{@{}c@{}}No\\ (0.3141)\end{tabular} &
  \textbf{\begin{tabular}[c]{@{}c@{}}Yes\\ (0.0001)\end{tabular}} &
  \textbf{\begin{tabular}[c]{@{}c@{}}Yes\\ (0.0001)\end{tabular}} &
  \textbf{\begin{tabular}[c]{@{}c@{}}Yes\\ (0.0001)\end{tabular}} &
  \textbf{\begin{tabular}[c]{@{}c@{}}Yes\\ (0.0003)\end{tabular}} &
  \textbf{\begin{tabular}[c]{@{}c@{}}Yes\\ (0.0001)\end{tabular}} &
  \textbf{\begin{tabular}[c]{@{}c@{}}Yes\\ (0.0001)\end{tabular}} \\
\begin{tabular}[c]{@{}c@{}}GPT-OSS-20B\\ (Advanced)\end{tabular} &
   &
  N/A &
  \textbf{\begin{tabular}[c]{@{}c@{}}Yes\\ (0.0002)\end{tabular}} &
  \textbf{\begin{tabular}[c]{@{}c@{}}Yes\\ (0.0001)\end{tabular}} &
  \textbf{\begin{tabular}[c]{@{}c@{}}Yes\\ (0.0005)\end{tabular}} &
  \textbf{\begin{tabular}[c]{@{}c@{}}Yes\\ (0.0001)\end{tabular}} &
  \textbf{\begin{tabular}[c]{@{}c@{}}Yes\\ (0.0001)\end{tabular}} &
  \textbf{\begin{tabular}[c]{@{}c@{}}Yes\\ (0.0001)\end{tabular}} \\
\begin{tabular}[c]{@{}c@{}}GPT-4o\\ (Base)\end{tabular} &
   &
   &
  N/A &
  \textbf{\begin{tabular}[c]{@{}c@{}}Yes\\ (0.0454)\end{tabular}} &
  \begin{tabular}[c]{@{}c@{}}No\\ (0.3839)\end{tabular} &
  \begin{tabular}[c]{@{}c@{}}No\\ (0.4432)\end{tabular} &
  \textbf{\begin{tabular}[c]{@{}c@{}}Yes\\ (0.0462)\end{tabular}} &
  \textbf{\begin{tabular}[c]{@{}c@{}}Yes\\ (0.0001)\end{tabular}} \\
\begin{tabular}[c]{@{}c@{}}GPT-OSS-20B\\ (Base)\end{tabular} &
   &
   &
   &
  N/A &
  \begin{tabular}[c]{@{}c@{}}No\\ (0.0833)\end{tabular} &
  \begin{tabular}[c]{@{}c@{}}No\\ (0.1024)\end{tabular} &
  \begin{tabular}[c]{@{}c@{}}No\\ (0.2433)\end{tabular} &
  \textbf{\begin{tabular}[c]{@{}c@{}}Yes\\ (0.0002)\end{tabular}} \\
SVM &
   &
   &
   &
   &
  N/A &
  \begin{tabular}[c]{@{}c@{}}No\\ (0.4567)\end{tabular} &
  \textbf{\begin{tabular}[c]{@{}c@{}}Yes\\ (0.0376)\end{tabular}} &
  \textbf{\begin{tabular}[c]{@{}c@{}}Yes\\ (0.0007)\end{tabular}} \\
LR &
   &
   &
   &
   &
   &
  N/A &
  \textbf{\begin{tabular}[c]{@{}c@{}}Yes\\ (0.0371)\end{tabular}} &
  \textbf{\begin{tabular}[c]{@{}c@{}}Yes\\ (0.0005)\end{tabular}} \\
Llama3-8B &
   &
   &
   &
   &
   &
   &
  N/A &
  \textbf{\begin{tabular}[c]{@{}c@{}}Yes\\ (0.0107)\end{tabular}} \\
Longformer &
   &
   &
   &
   &
   &
   &
   &
  N/A \\ \hline
\end{tabular}%
}
\end{table*}


\section{Discussion}

During prompt engineering using GPT-4o, we observed cases where the model exhibited a tendency to classify all candidate reports as non–follow-ups. Prompts that included overly detailed instructions often led to incorrect predictions, highlighting LLMs' sensitivity to prompt complexity. Even with optimized prompting, GPT-4o (Base) achieved an F1 score of 0.772 (95\% CI: 0.738–0.803), which was not statistically different from SVM (F1 = 0.777, 95\% CI: 0.743–0.809) or LR (F1 = 0.775, 95\% CI: 0.740–0.808). This finding suggests that effective prompt engineering is essential for GPT-4o, while more sophisticated, task-tailored approaches may still be required for it to consistently outperform simpler, more efficient traditional ML models.

The examples shown in Figure~\ref{aba:fig5} demonstrate the benefit of using task-specific, condensed input for GPT-4o (Advanced). The index report documents multiple findings, including \textit{``left upper lobe consolidation''} and \textit{``hypodensities within the spleen,''} but the follow-up recommendation specifically concerns the splenic hypodensities and calls for a CT abdomen and pelvis. The candidate report, however, is not a CT of the recommended anatomy and does not address the splenic findings, making it unqualified as an appropriate follow-up. In the Base setting, which provided the full index report as input, the model incorrectly predicted this case as a valid follow-up by focusing only on lexical overlap. In contrast, the Advanced setting—by restricting input to the recommendation sentence and metadata—correctly identified the mismatch in anatomy and modality. We anticipate that this targeted input design reduced false positives, leading to improved precision (0.740 vs. 0.841).

Extending this analysis, the open-source GPT-OSS-20B model provided additional insight into model-specific prompt sensitivity. When using the same prompt as GPT-4o, GPT-OSS-20B exhibited lower performance (F1 = 0.799, 95\% CI: 0.767–0.831), primarily due to over-strict reasoning that rejected valid follow-ups when not all findings from the recommendation were addressed. After refining the prompt to clarify that \textit{"a follow-up remains valid even if only a subset of the recommended findings is discussed"}, GPT-OSS-20B (Advanced) improved substantially to 0.828 F1 (95\% CI: 0.799–0.856), achieving performance statistically indistinguishable from GPT-4o (Advanced) (p = 0.3141). This demonstrates that small, conceptually meaningful adjustments to task framing can harmonize LLM reasoning with clinical logic, particularly for open-source models whose instruction-following behavior may differ from proprietary counterparts.

It is noteworthy that feature-based models such as LR and SVM performed comparable or better than GPT-4o (Base) and GPT-OSS-20B (Base), underscoring their value as interpretable and resource-efficient alternatives for deployment in clinical applications. These models are especially advantageous in settings where computational resources for model development, inference, and validation are limited. In our study, analysis of the LR feature weights revealed that the most influential predictors combined metadata (e.g., time gap, anatomy) with terms describing findings and their characteristics. Words such as ``\textit{nodule,''} ``\textit{lesion,''} \textit{``unremarkable,''} and \textit{``benign''} were among the strongest positive contributors—closely aligning with the reasoning processes employed by radiologists.


\section{Conclusion}
In this study, we present a newly annotated corpus of 6,393 radiology reports from 586 patients, designed to support the development and benchmarking of models for identifying imaging follow-ups. Using this resource, we comprehensively evaluated a spectrum of methods ranging from traditional feature-based classifiers and transformer-based encoders to recent generative LLMs. Among all evaluated systems, GPT-4o (Advanced) achieved the highest performance (F1 = 0.832), closely matching inter-annotator agreement (F1 = 0.846). The open-source GPT-OSS-20B (Advanced), when guided by a refined prompt incorporating task-specific clarification, achieved comparable results (F1 = 0.828) without a statistically significant difference.

These results underscore the importance of task-aware prompt design and input curation in optimizing LLM reasoning for clinical applications. While closed-source models such as GPT-4o demonstrate strong out-of-the-box performance, open-source systems like GPT-OSS-20B achieve comparable accuracy with greater flexibility for fine-tuning and integration within secure institutional environments. Despite the advances of LLMs, traditional models such as logistic regression (LR) and support vector machines (SVM) remain valuable, offering interpretable and computationally efficient baselines. Collectively, this corpus and evaluation provide a reproducible foundation for future research, highlighting a complementary landscape where open and closed LLMs, alongside interpretable classical models, together advance the robustness and transparency of clinical NLP systems.

\section{Limitations}

Accurate identification of follow-up imaging has important implications for clinical practice, including reducing unnecessary scans and improving the management of incidental findings. To enhance model generalizability, future work should leverage multi-institutional datasets encompassing a broader range of imaging modalities and clinical scenarios. Multi-modal approaches that integrate imaging data with radiology text, as well as the use of domain-adapted vision-language models (e.g., MedGemma \cite{sellergren2025medgemma}, MedVLM-R1 \cite{pan2025medvlm}), represent promising avenues for advancing performance and clinical relevance.

Finally, evaluating LLM outputs should extend beyond standard performance metrics such as precision, recall, and F1. Our current study was limited to these measures, but future evaluations should incorporate radiologist review, clinical reasoning assessments, and systematic evaluations of reasoning quality, factual consistency, and clinical validity. Establishing such robust evaluation frameworks will be essential to ensure the safe and effective deployment of LLMs in real-world healthcare settings.

\section{Ethics Statement}

We obtained approval from our institution’s Institutional Review Board (IRB) with a waiver of informed consent for the use of clinical text data. Radiology reports may contain patient Protected Health Information (PHI), including names, contact information, and other identifiers. To ensure patient privacy, all reports were automatically de-identified using a neural de-identification model, followed by manual review and secondary de-identification by trained medical student annotators to verify that no residual PHI remained. Both the original and de-identified reports were securely stored on a Health Insurance Portability and Accountability Act (HIPAA)-compliant server. All researchers and annotators completed human subjects training and were authorized to handle data containing PHI.

The annotated corpus used in this study was randomly sampled from the general population of patients undergoing imaging examinations at a single academic medical center. The dataset includes a diverse range of imaging modalities and clinical contexts, and was developed to support the task of identifying imaging follow-ups. Demographic variables were not used during sampling, and the patient population may not be representative of other institutions or the broader population. Differences in report style, structure, and documentation practices across sites may therefore affect the generalizability of the developed models.

\section{Acknowledgments}
This work was supported in part by the National Institutes of Health and the National Cancer Institute (NCI) (GrantNr. XXX). The content is solely the responsibility of the authors and does not necessarily represent the official views of the National Institutes of Health.


\section{Bibliographical References}\label{sec:reference}

\bibliographystyle{lrec2026-natbib}
\bibliography{lrec2026-example}

\bibliographystylelanguageresource{lrec2026-natbib}
\bibliographylanguageresource{languageresource}

\end{document}